\documentclass[10pt,twocolumn,letterpaper]{article}

\usepackage{iccv}
\usepackage{times}
\usepackage{epsfig}
\usepackage{graphicx}
\usepackage{amsmath}
\usepackage{amssymb}

\usepackage{booktabs}

\usepackage{xcolor} 

\usepackage[ruled]{algorithm2e}
\usepackage{multirow}

\usepackage[pagebackref=true,breaklinks=true,letterpaper=true,colorlinks,bookmarks=false]{hyperref}

\iccvfinalcopy 

\def\iccvPaperID{8891} 

\ificcvfinal\fi

\begin{document}

\title{TDG: Text-guided Domain Generalization}

\author{Geng Liu\\
Institute of Automation, \\
Chinese Academy of Sciences\\
{\tt\small  liugeng2020@ia.ac.cn}
\and
Yuxi Wang\\
Centre for Artificial Intelligence and Robotics, HKISI,\\
Chinese Academy of Sciences\\
{\tt\small yuxiwang93@gmail.com}
\and
Zhaoxiang Zhang\\
Institute of Automation, \\
Chinese Academy of Sciences\\
{\tt\small zhaoxiang.zhang@ia.ac.cn}
}

\maketitle
\ificcvfinal\thispagestyle{empty}\fi

\begin{abstract}
Domain generalization (DG) attempts to generalize a model trained on single or multiple source domains to the unseen target domain. 
Benefiting from the success of Visual-and-Language Pre-trained models in recent years, we argue that it is crucial for domain generalization by introducing extra text information. 
In this paper, we develop a novel \textbf{T}ext-guided \textbf{D}omain \textbf{G}eneralization (TDG) paradigm for domain generalization, which includes three following aspects. 
Specifically, we first devise an automatic words generation method to extend the description of current domains with novel domain-relevant words. Then, we embed the generated domain information into the text feature space, by the proposed prompt learning-based text feature generation method, which shares a common representation space with the image feature. Finally, we utilize both input image features and generated text features to train a specially designed classifier that generalizes well on unseen target domains, while the image encoder is also updated under the supervision of gradients back propagated from the classifier.
Our experimental results show that the techniques incorporated by TDG contribute to the performance in an easy implementation manner. Experimental results on several domain generalization benchmarks show that our proposed framework achieves superior performance by effectively leveraging generated text information in domain generalization. 
\end{abstract}

\begin{figure}[!tp]
    \centering
    \includegraphics[width=1.0\linewidth]{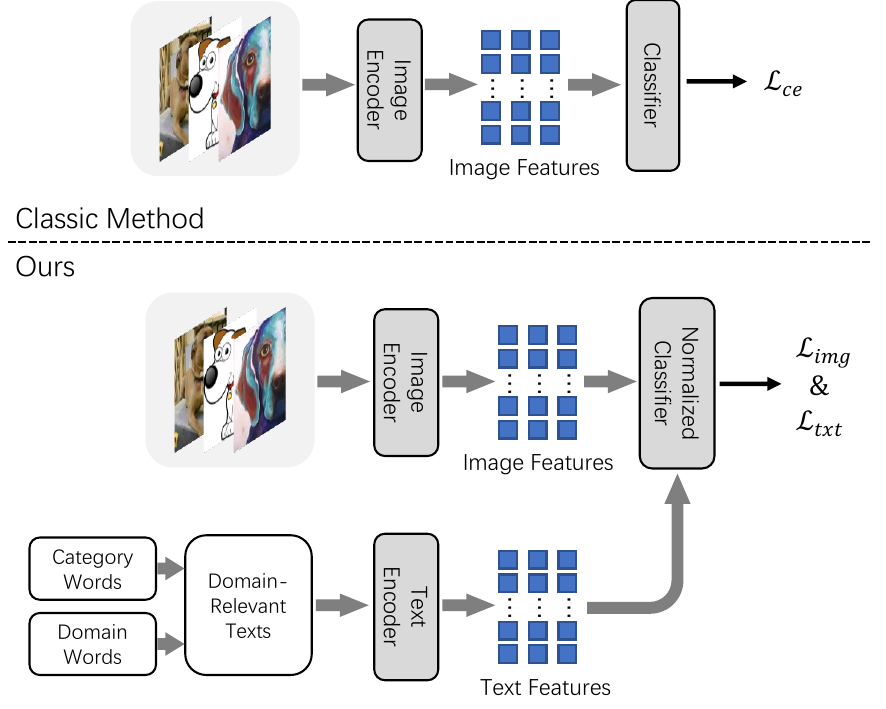}
    \caption{Difference between classic DG methods and ours. Classic methods concentrate on learning generalizability from input images over source domains. On the contrary, our method leverages extra domain information in generated domain-relevant texts to further improve the performance of models.
}
    \label{fig:Compare}
\vspace{-8pt}
\end{figure}

\section{Introduction}
\label{sec:intro}
    Machine learning has gained great development these years in many fields, such as computer vision and natural language processing. The goal of traditional machine learning is that, given a batch of training data with annotations, design a model and train it on labeled data to enable the model to do predictions on unseen test data. Ordinarily, if the distributions of training data and test data are similar, for example, they are all photos taken in the same scene, the model trained on training data will perform well on test data. However, in practice, the test data is usually collected from different scenarios performing with significant degradation, which is known as domain shift. On the other hand, collecting and annotating all datasets are not always available as they are labor-expensive and time-consuming. Alternatively, existing researches \cite{dou2019domain,muandet2013domain,Volpi} attempt to struggle for the generalization model that aims to improve the robustness of the network on various unseen test domains, which is important for tackling domain shift problem, especially for out-of-distribution (OOD) scenarios. The model has a fair ability to do predictions on test data. However, with the widening of the distribution gap between training data and test data, the performance of the model tends to be worse. Then the concept of domain generalization (DG) was proposed to address this problem by strengthening the generalization of models.

Generally speaking, domain generalization models are trained on images from a single or multiple related but distinct source domains, and then are tested on another unseen target domain. The distribution of source and target domains varies widely, but categories of them are the same. Plenty of domain generalization methods are proposed these years, which can be roughly divided into three broad categories: augmentation-based DG \cite{du2020learning, motiian2017unified,Volpi,li2021simple}, alignment-based DG \cite{JiGen,RSC,wang2018learning,xu2020robust,zhou2020learning}, and meta-learning-based DG \cite{balaji2018metareg,li2018learning,zhao2021learning,finn2017model}. Some of which focus on manipulating data, and some of which focus on developing better learning algorithms and loss functions. 
However, experimental results show that the performances of most proposed domain generalization methods are limited, which can hardly surpass the performance of trivial baseline ERM \cite{ERM} remarkably under fair comparison.

Traditional DG methods focus on learning generalizability from source images, as Fig. \ref{fig:Compare} shows, while the improvement has come to a bottleneck. Therefore, some new methods focus on leveraging extra information beyond training datasets into training or testing processes.
Utilizing multi-modal information such as text information in DG tasks seems to be reasonable for texts could be edited to contain any kind of semantic information easily.
But in general, text information and image information extracted by different extractors have their own feature spaces, which are not the same as each other, thus we can not just utilize text information in image tasks directly. Fortunately, benefiting from large-scale vision-language pre-trained model \cite{CLIP}, we can explore the domain information hidden in the text embedding, with the common feature space between visual and textual inputs. 

In this paper, we propose Text-guided Domain Generalization (TDG), an approach that utilizes texts to improve the performance of models on DG tasks. We utilize a pre-trained lexical substitution model to automatically generate diverse domain-relevant words, which contain abundant domain information. Then we design a prompt learning module to train a text prompt template, which will be filled with category words and domain words to generate texts containing corresponding category information and domain information. Finally, text features extracted from generated texts are treated as augmentation of image features and are simply used to train a specially designed normalized classifier, which is used to do predictions cooperating with the image encoder. Different from existing VLP-based methods \cite{coop,cocoop,DPL,DUPR} that concentrate on text-and-image feature pair alignment by prompt tuning, our TGD explicitly explores the domain information in the text by diversifying the textual inputs and then pulls the image embedding into the text embedding for handling out-of-distribution problem. 

The contributions of this paper are summarized as follows:
\begin{enumerate}
    \item We propose a lexical substitution model-based method to automatically generate various domain-relevant words.
    \item We propose a prompt learning-based method to train the text prompt template, thus promoting the domain diversity of generated texts.
    \item We conduct extensive experiments on several domain generalization benchmarks and achieve state-of-the-art performance on DG tasks, demonstrating the effectiveness of the proposed framework.
\end{enumerate}
\begin{figure*}[!tp]
    \centering
    \includegraphics[width=1.0\linewidth]{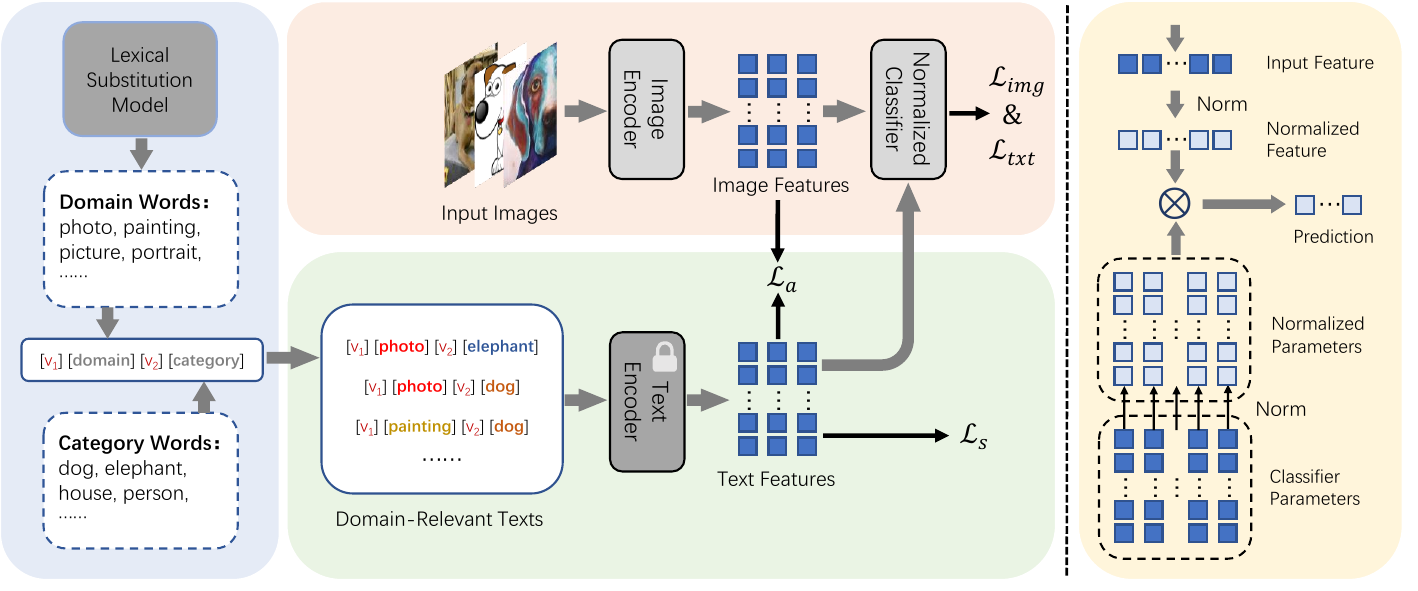}
    \caption{\textbf{Left}: Overview of our method. Our method mainly contains three components, namely domain-relevant words generation module, text generation module, and normalized classifier. Firstly, we utilize a pre-trained lexical substitution model to generate a lot of words that are relevant to the domain of images, such as photo, painting, etc. Then we fill a pre-designed text prompt template with category words in the current dataset and generated domain-relevant words to create texts containing corresponding category information and domain information. After that we update the text prompt template by maximizing similarities between text features and image features that are in the same category; while minimizing similarities between text features themselves which contain the same category word but different domain words. Finally, we utilize both input image features and generated text features to train the normalized classifier and the image encoder under the supervision of cross-entropy loss.
    \textbf{Right}: Illustration of normalized classifier. In the classification process, both the input feature and parameters of every classification head are normalized to unit vectors and are multiplied to calculate the output prediction.
}
\label{fig2}
\vspace{-8pt}
\end{figure*}

\section{Related Work}
\subsection{Domain Generalization}

We list some of the main-stream domain generalization methods below, which are divided into two categories according to their key point. Data augmentation methods focus on augmenting the training set by adding novel images, while representation learning methods focus on learning to extract better features.

\noindent\textbf{Data Augmentation} is widely used in image tasks to improve the robustness of models, it is also effective in DG tasks. Classical data augmentation operations such as random horizontal flips, random color jitter, grayscaling, etc, perform well in DG tasks \cite{DomainBed}.
Volpi \etal \cite{Volpi} augments the training set by producing new images that are hard to classify for the current model, thus the model learns to classify the worst case during training.
DDAIG \cite{DDAIG} proposes an adversarial domain transformation network to transform original images into unseen domains.
L2A-OT \cite{L2A-OT} utilizes StarGAN \cite{StarGAN} to generate images in novel domains, which maximizes the distance between original images and augmented images while maintaining semantic consistency.
AugMix \cite{AugMix} randomly samples data augmentation operations and corresponding weights to compose a mixed operation, which produces new images without veering too far from the original.
FACT \cite{FACT} transforms input images into amplitude components and phase components using Fourier transformation, then augments images by mixing the amplitude components from different images while keeping phase components unchanged and reconstruct images using the combination of both components.

\noindent\textbf{Representation Learning} is the objective of machine learning, many DG methods improve the generalizability of models by learning domain-invariant representations.
Motiian \etal \cite{motiian2017unified} proposed semantic alignment loss and separation loss. Semantic alignment loss minimizes the distance between images in the same class over different domains, and separation loss maximizes the distance between images in different classes over different domains.
Tzeng \etal \cite{MMD} minimizes the feature distribution divergence by minimizing the maximum mean discrepancy (MMD).
Li \etal \cite{li2018domain} and Peng \etal \cite{peng2019moment} improve the robustness of features by aligning features extracted from images in the same category.
Ganin \etal \cite{DANN} firstly leverages adversarial learning into domain adaptation. They add a domain classifier connected to the feature extractor via a gradient reversal layer, which multiplies the gradient by a negative constant during back-propagation. The domain classifier is trained to classify domains correctly, at the same time the feature extractor is trained to make the features hard to classify in the aspect of the domain due to the gradient reversal layer. 
Similar to DANN, Li \etal \cite{CIAN} proposes a conditional invariant adversarial network to learn class-conditional adversarial networks.

\subsection{Vision and Language Pre-training}
CLIP \cite{CLIP} is a big vision-language model trained on 400 million image-text pairs across contrastive learning, and brings great improvement on many downstream tasks. Thanks to the huge training set, CLIP performs impressively great on zero-shot tasks without any fine-tuning. CLIP is able to project images and texts to the same feature space, in which the image and the text semantically similar are also similar. Based on such characteristic, CoOp \cite{coop} firstly leverage prompt learning into vision-language model fine-tuning, which is already widely used in NLP tasks. CoOp makes significant progress on few-shot image classification tasks. However, a disadvantage of CoOp is that invariant prompt context may over-fits the training dataset, and its performance is limited on wider unseen datasets. To address this problem, CoCoOp \cite{cocoop} further improves CoOp by learning a simple meta-net to generate image-wise bias for prompt context, thus getting stronger generalizability.


\section{Method}
In the typical context of domain generalization (DG), we assume to have access to similar but distinct source domains $\mathcal{D}_{S} = \{\mathcal{D}_1, \mathcal{D}_2, \dots, \mathcal{D}_{N_D} \}$, with $n_k$ labeled samples $\{(x_i^k, y_i^k)\}_{i=1}^{n_k}$ in the $k$-th source domain $\mathcal{D}_k$. $N_D$ is the number of source domains. $(x_i^k, y_i^k)$ denotes the pair of visual images and the corresponding semantic labels, and $y_i^k \in \{1, 2, \dots, N_c\}$, $N_c$ is the number of categories. The target domain is denoted by $\mathcal{D}_t=\{x_{i}^{t}\}_{i=1}^{n_t}$ with no labeled information. $n_t$ is the number of target images in $\mathcal{D}_t$. Note that the target domain is normally different from the source domains but has identical label space with the source domains. The goal of domain generalization is to learn a domain-agnostic model $f(\cdot; \theta)$ on the source domains that expect to perform well on the unseen target domain $\mathcal{D}_t$. 

To tackle this problem, unlike prior works which purely concentrate on making better use of training images from different source domains, we access extra domain information from elaborately generated texts. We embed abundant domain information into texts and extract text features from these texts through the CLIP text encoder. For features extracted by the image encoder and text encoder to be in the same feature space, we could simply utilize text features to train the classifier as well as image features. Therefore, domain information contained in text features could improve the generalizability of the classifier in a direct manner. The image encoder is also trained to extract features that could generalize well under the supervision of gradients back propagated from the classifier.
Overview of our method is shown in Figure~\ref{fig2}, and below we introduce the components of our method in detail.

\subsection{Domain-Relevant Words Generation}
To begin with, as we want to utilize texts to improve the performance of models on unseen target domains, it is necessary to generate texts that are semantically relevant to the domain of images, such as photo, painting, etc. To accomplish this purpose, we design a domain-relevant words generation module based on the pre-trained lexical substitution model proposed by Nikolay \etal \cite{LexSub}. The lexical substitution model is trained to substitute any word of the given sentence with different but semantically reasonable words, thus it is easy to generate a number of domain-relevant words using the model. In practice, we design a simple original sentence ``an image of a dog" and input it into the lexical substitution model with the second word ``image" set to be substituted, then the model returns a list of candidate words. Generated candidate words are shown in Table~\ref{table1}, as we can see, most of the generated words are relative to the domain of images and are semantically different from each other. 

\begin{table}[]
\renewcommand\arraystretch{1.2}
\centering
\caption{Domain-relevant words generated by pre-trained lexical substitution model. We simply input an original text shown in table to the lexical substitution model and set ``image" to be substituted, thus the lexical substitution model returns a number of words that could be used to substitute ``image" to constitute reasonable texts.}
\vspace{4pt}
\begin{tabular}{c}
\hline\noalign{\smallskip}
\multicolumn{1}{c}{\textbf{Original text:} ``An [\textcolor{gray}{image}] of a dog".}\\
\noalign{\smallskip}\hline\noalign{\smallskip}
\multicolumn{1}{c}{\textbf{Generated words:} ``picture", ``photo", ``photograph",}\\
\multicolumn{1}{c}{``portrait",``silhouette", ``statue", ``symbol", ``painting",}\\
\multicolumn{1}{c}{``figure", ``depiction", ``drawing", ``caricature", ``video",}\\
\multicolumn{1}{c}{``face", ``sculpture", ``vision", ``illustration”, ``cartoon”,}\\
\multicolumn{1}{c}{``imagery”, ``representation".}\\ \noalign{\smallskip}\hline
\end{tabular}
\vspace{-8pt}
\label{table1}
\end{table}

\begin{figure}[!tp]
    \centering
    \includegraphics[width=0.8\linewidth]{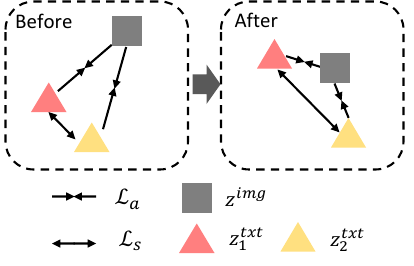}
    \caption{Simple illustration of proposed loss functions $\mathcal{L}_{a}$ and $\mathcal{L}_{s}$. $z^{img}$ is feature of image in specific category; $z^{txt}_1$ and $z^{txt}_2$ are features of texts containing same category word as $z^{img}$, but containing different domain words. $\mathcal{L}_{a}$ reduces the distance between $z^{txt}_1$, $z^{txt}_2$ and $z^{img}$; $\mathcal{L}_{s}$ enlarges the distance between $z^{txt}_1$ and $z^{txt}_2$. By minimizing $\mathcal{L}_{a}$ and $\mathcal{L}_{s}$, text features contain more diverse domain information while keeping category information unspoiled, thus are more helpful to domain generalization tasks.
}
\label{fig3}
\vspace{-4pt}
\end{figure}

\subsection{Text Generation via Prompt Learning} 
\label{Text Generation via Prompt Learning}
With the module proposed above, we get a bunch of domain-relevant words in hand, but these words cannot be used in the training process directly.
In this section, we construct a prompt learning module to train a text prompt template:``[\textcolor{red}{$v_1$}] [\textcolor{gray}{domain}] [\textcolor{red}{$v_2$}] [\textcolor{gray}{category}]", in which [\textcolor{red}{$v_1$}] and [\textcolor{red}{$v_2$}] are learnable embedding, [\textcolor{gray}{domain}] is embedding of any domain word and [\textcolor{gray}{category}] is embedding of any category word. We could generate texts relevant to variant domains and categories by filling this template with category words of the dataset and domain-relevant words, and extract text features from generated texts using text encoder.

Obviously, we wish that extracted text features are helpful to DG tasks, but what kind of text feature is more helpful? In our opinion, when we fill the text prompt with different domain words and category words to generate texts, text features extracted from texts should be representative of the corresponding category of images and contain very different domain information. Text features with such characteristics could provide more novel domain information during training and are helpful for generalization. To address this issue, we propose two loss functions to constrain the learnable embedding of the text prompt template. 
Firstly, category information in text features should be kept in the training process, therefore we align text features with image features in the same category to maintain the category information of text features unspoiled through the loss function below:
\begin{equation}
   \mathcal{L}_{a}=-\frac{1}{N_{s}}\frac{1}{N_{d}}\sum_{n=1}^{N_s}\sum_{i=1}^{N_c}\sum_{j=1}^{N_d}\alpha_{ni}\langle z_n^{img},z_{ij}^{txt}\rangle.
\end{equation}

In this equation, $\langle \cdot\rangle$ is cosine similarity; $N_s$ denotes the number of images in source domains; $N_d$ denotes the number of generated domain-relevant words.
We use $E_{img}(\cdot)$ and $E_{txt}(\cdot)$ to denote the image encoder and the text encoder respectively, then $z_n^{img}=E_{img}(x_n^s)$ denotes features extracted by image encoder from source image $x_n^s$, $z_{ij}^{txt}=E_{txt}(t_{ij})$ denotes features extracted by text encoder from text $t_{ij}$, which consists of the $i$-th category word and the $j$-th domain-relevant word. 
$\alpha_{ni}=1$ if and only if $x_n^s$ belongs to the $i$-th class, otherwise $\alpha_{ni}=0$. While the similarity loss above is minimized, similarities between image features and text features in the same category are maximized.

Then, we enlarge the differences between text features, which are extracted from texts containing the same category words but different domain words through the loss function below:
\begin{equation}
   \mathcal{L}_{s}=\frac{1}{N_{c}}\frac{1}{N_{d}^2}\sum_{i=1}^{N_c}\sum_{j,j'=1}^{N_d}\langle z_{ij}^{txt},z_{ij'}^{txt}\rangle.
\end{equation}
By minimizing such similarity loss, text features are encouraged to be more sensitive to the change of the domain word in texts, thus forcing text features to be more diverse in the aspect of the domain. 
Total loss of text prompt learning is: 
\begin{equation}
   \mathcal{L}_{pl}=\mathcal{L}_{a}+\lambda\mathcal{L}_{s},
\end{equation}
in which $\lambda$ is the weight of similarity loss. As shown in Figure~\ref{fig3}, by minimizing the two loss functions above with appropriate $\lambda$ in the prompt learning process, we are able to get a bunch of text features containing representative category information and diverse domain information, which benefits the generalizability of models greatly. We set $\lambda$ to $0.3$ in all of the experiments.

\subsection{Normalized Classifier}
In traditional ResNet-based DG methods, linear classifiers and MLP classifiers are widely used, which perform well in most of the methods. 
Since a lot of VLP models (such as CLIP, ALIGN \cite{ALIGN}, CoCa \cite{CoCa}) are pre-trained to align image features and text features into the same feature space, adopting cosine similarities between image features and text features as the constraint during training. The orientations of features in feature space are valuable information in features, while the scale of every single logit is invalid information. Based on such characteristics, we specially design a normalized classifier for VLP models to make the classification process similar to their pre-training process. 

Detailed illustration of the normalized classifier is shown in Figure~\ref{fig2}. \textbf{Right.} Input features are firstly normalized to unit vector: $\hat{z}=Norm(z)$.
Given a linear classifier $G$, we use $g_i$ to denote parameters of the $i$-th classification head corresponding to the $i$-th category.
Similar to input features, parameters of each classification head are normalized to the unit, too: $\hat{g_i}=Norm(g_i)$, then normalized classification heads constitute the normalized classifier $\hat{G}=\{g_i\}_{i=1}^{N_c}$. Finally, the prediction of the input feature $z$ is $\hat{y}=\hat{G}(z)$.
Compared with the linear classifier which directly multiplies input features with parameters of classification heads to output prediction logits, the normalized classifier ignores the invalid information in input features and thus is more robust in dealing with features from CLIP.

\begin{algorithm}[!t]
\small
\caption{Text-guided Domain Generalization}
\KwIn{Source domains $\mathcal{D}_{S} = \{\mathcal{D}_1, \dots, \mathcal{D}_{N_D} \}$, domain-relevant words $W_d=\{d_1,..., d_{N_d}\}$, category words of dataset $W_c=\{c_1,..., c_{N_c}\}$, text prompt template $T(\cdot,\cdot)$, CLIP image encoder $E_{img}(\cdot)$, CLIP text encoder $E_{txt}(\cdot)$, normalized classifier $\hat{G}$.} 
\KwOut{Learned classifier $\hat{G}$}.
\For{sampled mini-batch $\{(x_n^s,y_n^s)\}_{n=1}^{B}\in\mathcal{D}_{S}$}{
    \For{$0\leq n<B$}{
        $z_n^{img}=E_{img}(x_n^s)$\;
        $\hat{y}_n^{img}=\hat{G}(z_n^{img})$\;
    }
    \For{category word $c_i\in W_c$}{
        \For{domain-relevant word $d_j\in W_d$}{
            $t_{ij}=T(c_i,d_j)$\;
            $z_{ij}^{txt}=E_{txt}(t_{ij})$\;
        }
    } 
    $\mathcal{L}_a=-\frac{1}{N_{s}}\frac{1}{N_{d}}\sum_{n=1}^{B}\sum_{i=1}^{N_d}\sum_{j=1}^{N_d}\alpha_{ni}\langle z_n^{img},z_{ij}^{txt}\rangle$\;
    $\mathcal{L}_s=\frac{1}{N_{c}}\frac{1}{N_{d}^2}\sum_{i=1}^{N_c}\sum_{j,j'=1}^{N_d}\langle z_{ij}^{txt},z_{ij'}^{txt}\rangle$\;
    update $T(\cdot,\cdot)$ to minimize $\mathcal{L}_a+\lambda \mathcal{L}_s$\;
    \For{category word $c_i\in W_c$}{
        \For{domain-relevant word $d_j\in W_d$}{
            $t_{ij}=T(c_i,d_j)$\;
            $z_{ij}^{txt}=E_{txt}(t_{ij})$\;
            $\hat{y}_{ij}^{txt}=\hat{G}(z_{ij}^{txt})$\;
        }
    }
    $\mathcal{L}_{img}=\frac{1}{B}\sum_{n=1}^{B}CrossEntropy(\hat{y}_n^{img},y_n^s)$\;
    $\mathcal{L}_{txt}=\frac{1}{N_{c}}\frac{1}{N_{d}}\sum_{i=1}^{N_c}\sum_{j=1}^{N_d}CrossEntropy(\hat{y}_{ij}^{txt},y_i^{txt})$\;
    \If {warm up}{update $\hat{G}$ to minimize $\mathcal{L}_{img}$+$\mathcal{L}_{txt}$}
    \Else {
        update $\hat{G}$ and $E_{img}$ to minimize $\mathcal{L}_{img}$+$\mathcal{L}_{txt}$
    }
}
\end{algorithm}

Finally, we update the normalized classifier $\hat{G}$ and the image encoder $E_{img}$ utilizing both image features and text features under the supervision of cross-entropy loss functions:
\begin{equation}
   \mathcal{L}_{img}=-\frac{1}{N_{s}}\sum_{n=1}^{N_{s}}y_{n}^{s}\log \hat{y}_{n}^{img},
\end{equation}
\begin{equation}
   \mathcal{L}_{txt}=-\frac{1}{N_{c}}\frac{1}{N_{d}}\sum_{i=1}^{N_{c}}\sum_{j=1}^{N_{d}}y_{i}^{txt}\log \hat{y}_{ij}^{txt},
\end{equation}
\begin{equation}
   \mathcal{L}_{ce}=\mathcal{L}_{img}+\mathcal{L}_{txt},
\end{equation}
in which $y_{i}^{txt}$ is the label of the $i$-th category.
The training procedures are detailed in Algorithm 1.


\setlength{\tabcolsep}{2.5pt}
\begin{table*}[ht]
\renewcommand\arraystretch{1.3}
\centering
\caption{Comparison of performance between different methods on PACS, OfficeHome, VLCS, and TerraIncognita datasets. We use the bold font to highlight the best results on every domain and dataset. The listed methods are all based on ViT-B/16 backbone pre-trained by CLIP. 
}
\vspace{4pt}
\resizebox{\textwidth}{!}{
    \begin{tabular}{l|ccccc|ccccc|ccccc|ccccc}
    \hline
    \multirow{2}{*}{Method} & \multicolumn{5}{c|}{PACS} & \multicolumn{5}{c|}{OfficeHome} & \multicolumn{5}{c|}{VLCS} & \multicolumn{5}{c}{TerraIncognita}\\ \cline{2-21}
    \multicolumn{1}{c|}{} & A & C & P & S & \multicolumn{1}{l|}{Avg.} & A & C & P & R & \multicolumn{1}{l|}{Avg.} & C & L & S & V & \multicolumn{1}{l|}{Avg.} & L100 & L38 & L43 & L46 & Avg. \\ \hline
    SEDGE+\cite{SEDGE} & - & - & - & - & 96.10 & - & - & - & - & 80.70 & - & - & - & - & 82.20 &- & - & - & - & 56.80\\
    MIRO \cite{MIRO} & - & - & - & - & 95.60 &  - & - & - & - & 82.50 &  - & - & - & - & 82.20 &- & - & - & - & 54.30 \\
    DPL* \cite{DPL}& -&-&-&- & \textbf{97.30} &-&-&-&- & 84.20 & -&-&-&- & \textbf{84.30}  & -&-&-&- & 52.60 \\
    DUPR* \cite{DUPR}& -&-&-&- & 97.10 & -&-&-&- & 83.60 & -&-&-&- & 83.90  & -&-&-&- & 42.00\\
    ERM*\cite{ERM} & 94.34 & 98.38 & \textbf{99.88} & 90.12 & 95.68 & 80.51 & 70.79 & 88.83 & 89.08 & 82.30 & 99.72 & 58.77 & 77.57 & 84.86 & 80.23 & 47.23 & 56.08 & 51.79 & 48.04 & 50.78 \\
    ERM\cite{ERM} & 97.31 & 97.78 & 99.76 & 87.27 & 95.53 & 85.91 & 76.58 & 91.17 & 90.59 & 86.06 & 96.82 & 66.71 & 82.41 & 83.80 & 82.44 & \textbf{68.74} & 53.70 & 66.93 & 49.20 & 59.64 \\
    CoOp \cite{coop} & \textbf{98.54} & 98.55 & 99.82 & 89.67 & 96.65  & 82.12 & 73.31 & 90.29 & \textbf{91.48} & 83.06 & 99.51 & 60.09 & 78.70 & 85.40 & 80.92  & 48.53 & 52.67 & 51.74 & 48.46 & 50.35 \\
    CoCoOp \cite{cocoop}& 97.46 & \textbf{98.93} & 99.76 & 89.66 & 96.45 & 82.20 & 71.78 & 91.26 & 90.70 & 83.99 & \textbf{100.00} & 65.36 & 78.15 & \textbf{85.66} & 82.29  & 51.13 & 49.83 & 54.41 & 47.54 & 50.73 \\
    \textbf{TDG(Ours)} & 97.90 & 98.21& 99.70 & \textbf{90.84} & 96.66 &  \textbf{86.90} & \textbf{77.41} & \textbf{91.78} & 91.42 & \textbf{86.88} & 97.46 & \textbf{69.35} & \textbf{83.33} & 84.48 & 83.66 & 67.22 & \textbf{54.37} & \textbf{68.01} & \textbf{51.57} & \textbf{60.29}\\ 
    \hline
    \end{tabular}
}
\vspace{-8pt}
\label{table2}
\end{table*}
\section{Experiment}
In this section, we demonstrate the superiority of our method on several standard DG benchmarks. We also conduct extensive ablation studies to analyze the impact of proposed components.

\subsection{Datasets}
To evaluate the efficacy of the proposed method, we conduct extensive comparative evaluations on four DG benchmarks: \textbf{PACS} \cite{PACS}, \textbf{OfficeHome} \cite{OfficeHome}, \textbf{VLCS} \cite{VLCS}, and \textbf{TerraIncognita} \cite{TerraIncognita}. The detailed illustrations of these benchmarks are as follows.

\noindent\textbf{PACS} consists of 9,991 images with seven categories (\textit{dog, elephant, giraffe, guitar, horse, house, and person}) for training and testing. These images are depicted in four different domains, including \textit{Photo (\textbf{P}), Art (\textbf{A}), Cartoon (\textbf{C}), and Sketch (\textbf{S})}, presenting large domain discrepancy.

\noindent\textbf{VLCS} is an object recognition dataset that contains 10,792 images with five classes (\textit{bird, car, chair, dog, and person}). These images are collected from four photo datasets (\textit{Caltech101 (\textbf{C}), LabelMe (\textbf{L}), SUN09 (\textbf{S}), VOC2007 (\textbf{V})}), which have domain shift because the captured scenes vary from urban to rural, and the viewpoints are biased toward either side-views or non-canonical views.

\noindent\textbf{OfficeHome} is a large-scale domain adaptation benchmark that contains 65 classes of training images. It consists of 15,500 images totally from four domains (\textit{Art (\textbf{A}), Clipart (\textbf{C}), Product (\textbf{P}), Real-Word (\textbf{R})}). The domain shift mainly comes from image styles and viewpoints, which is much smaller than PACS. 

\noindent\textbf{TerraIncognita} contains 24,788 photos of wild animals in 10 classes (\textit{bird, bobcat, cat, coyote, dog, empty, opossum, rabbit, raccoon, and squirrel}), which are taken from four different locations (\textit{\textbf{L100}, \textbf{L38}, \textbf{L43}, \textbf{L46}}), differences between images in different domains are the scene of photos and size of objects.

\setlength{\tabcolsep}{4pt}
\begin{table*}[ht]
\renewcommand\arraystretch{1.3}
\small
\begin{center}
\caption{Ablation study of proposed modules.  \textbf{+Text} indicates leveraging text features extracted from generated texts into training process; \textbf{+Norm} refers to further substituting the linear classifier in ERM to the normalized classifier.
}
\label{table3}
\vspace{4pt}
\begin{tabular}{l|cccccc|cccccc}
\hline
\multicolumn{1}{c|}{\multirow{2}{*}{Methods}} & \multicolumn{6}{c|}{PACS} & \multicolumn{6}{c}{OfficeHome}\\ \cline{2-13}
\multicolumn{1}{c|}{} & A & C & P & S & Avg. & \multicolumn{1}{l|}{Gain} & A & C & P & R & Avg. & \multicolumn{1}{l}{Gain}\\ 
\hline
ERM & 97.31 & 97.78 & 99.76 & 87.27 & 95.53 & - & 84.47 & 75.23 & 90.52 & 89.97 & 85.05 & - \\
\textbf{+Text} & 96.97 & 97.91 & 99.82 & 90.53 & 96.31 & \textcolor[rgb]{0,0.8,0}{$+0.55$} & 85.91 & 76.58 & 91.17 & 90.59 & 86.06 & \textcolor[rgb]{0,0.8,0}{$+1.01$}\\
\textbf{+Norm} & 97.90 & 98.21& 99.70 & 90.84 & 96.66 & \textcolor[rgb]{0,0.8,0}{$+1.13$} & 86.90 & 77.41 & 91.78 & 91.42 & 86.88 & \textcolor[rgb]{0,0.8,0}{$+1.83$} \\ 
\hline
\multicolumn{1}{c|}{\multirow{2}{*}{Methods}} & \multicolumn{6}{c|}{VLCS} & \multicolumn{6}{c}{TerraIncognita}\\
\cline{2-13}
& C & L & S & V & Avg. & \multicolumn{1}{l|}{Gain} & L100 & L38 & L43 & L46 & Avg. & Gain \\ \hline
ERM           & 96.82 & 66.71 & 82.41 & 83.80 & 82.44 & - & 68.74 & 53.70 & 66.93 & 49.20 & 59.64 & -\\
\textbf{+Text} & 96.04 & 67.85 & 83.97 & 84.83 & 83.17 & \textcolor[rgb]{0,0.8,0}{$+0.73$} & 66.37 & 56.14 & 67.28 & 50.28 & 60.02 & \textcolor[rgb]{0,0.8,0}{$+0.38$}\\
\textbf{+Norm} & 97.46 & 69.35 & 83.33 & 84.48 & 83.66 & \textcolor[rgb]{0,0.8,0}{$+1.22$} & 67.22 & 54.37 & 68.01 & 51.57 & 60.29 & \textcolor[rgb]{0,0.8,0}{$+0.65$}\\
\hline
\end{tabular}
\vspace{-16pt}
\end{center}
\end{table*}

\subsection{Implement Details}
Our implementation is based on \textbf{DomainBed} \cite{DomainBed}, which is a widely used platform for DG. We utilize the default data augmentation provided in DomainBed during training, including random horizontal flips, random color jitter, and grayscaling. We also utilize exponential moving average during training to improve the generalizability of models.
We use Adam optimizer \cite{Adam} to train the classifier and the image encoder, and the learning rate is set to $1e$-$3$ and $5e$-$6$ respectively. 
We use SGD optimizer \cite{SGD} to train the text prompt template following the setting in CoOp \cite{coop}, and the learning rate is set to $1e$-$3$. We trained our model for $3000$ steps with batch size $32$ on all of the datasets. Note that we set the first $300$ steps as warm up steps, during which the image encoder is frozen and only the classifier is updated to avoid the corruption of image encoder caused by the randomly initialized classifier.

During constructing experiments, we follow prior works \cite{JiGen,li2017deeper,li2019episodic} applying leave-one-domain-out protocol for fair comparisons, where one domain is chosen as the unseen target domain and the remaining domains are used as source domains for training. For OfficeHome and PACS datasets, we use the data splits provided by Shu \etal \cite{shu2021open} to split images into the training set and validation set; For VLCS and TerraIncognita datasets, we randomly split each domain into 90\% for training and 10\% for validation. We train our model on the training splits and select the best model on the validation splits of all source domains. For testing, we evaluate the selected model on all images of the target domain.

\subsection{Quantitative Evaluation}
In this section, we evaluate the proposed method on several domain generalization benchmarks and present the comparisons with existing CLIP-based approaches, including ERM \cite{ERM}, ERM with frozen backbone (noted as ERM*), SEDGE+ \cite{SEDGE}, MIRO \cite{MIRO}, DUPR \cite{DUPR}, CoOp \cite{coop}, CoCoOp \cite{cocoop} and DPL \cite{DPL}. For a fair comparison, we have reproduced the results of ERM based on ViT backbone pre-trained by CLIP \cite{CLIP}. MIRO is a previous domain generalization work adopting CLIP image encoder as the backbone; CoOp, CoCoOp and DPL are related to our work aiming to fine-tune the vision and language pre-trained model through prompt learning; DUPR utilizes domain words to yield domain-unified representations, thus improve the generalizability of models. The corresponding results are shown in Table~\ref{table2}. Note that DPL and DUPR are followed with *, because we are not able to reproduce the results of DPL through released code, and the text pool of DUPR involves the target domain words, which is undesirable in the DG tasks.

\noindent\textbf{Results on PACS:} CLIP image encoder performs impressively well on PACS dataset, especially on cartoon domain and photo domain, accuracy on which are close to $100\%$. The mean accuracy of ERM on PACS dataset exceeds $95\%$. Although the baseline is strong enough, our method still improves its performance for $3.57\%$ on sketch domain, $0.59\%$ on art painting domain and $1.13\%$ on average. Performance of our method is competitive to other methods which adopt CLIP image encoder as the backbone.

\noindent\textbf{Results on OfficeHome:} Our method improves the performance of baseline for an average of $1.83\%$ on 4 domains. Our method outperforms all of the compared methods, surpasses MIRO for $4.38\%$, surpasses CoOp for $3.82\%$, and surpasses CoCoOp for $2.89\%$. 

\noindent\textbf{Results on VLCS:} Note that all of the images in VLCS dataset are photos, differences between images in different domains are the size of objects and the scene of photos. With diverse words, there is no remarkable style difference between images in different domains. Our method still improves the performance of the baseline for $1.22\%$, and achieves competitive performance compared with other ViT-based methods.

\noindent\textbf{Results on TerraIncognita:} Similar to VLCS dataset, all of the images in TerraIncognita dataset are photos, too. Differences between images in different domains are the location of cameras. Our method improves the performance of baseline for $0.65\%$, and surpasses other ViT-based methods by a large margin.

\subsection{Ablation Study}
\noindent\textbf{Influence of Proposed Modules}
\label{Ablation Study of Proposed Modules}
We investigate the impact of the normalized classifier and the text generation module on 4 datasets, results are shown in Table~\ref{table3}. 
The baseline of our method is ERM with linear classifier. 
As we can see, after leveraging texts generated by our method into the training process, the performances of models are lifted significantly, proving the effectiveness of texts containing domain information. The normalized classifier also outperforms the linear classifier on all of the datasets. These experimental results demonstrate that the normalized classifier is more suitable for CLIP image encoder on DG tasks compared to the linear classifier.

\setlength{\tabcolsep}{4pt}
\begin{table}[!t]
\renewcommand\arraystretch{1.2}
\small%
\begin{center}
\caption{Ablation study of the prompt learning module on OfficeHome dataset. \textbf{Text} refers to utilizing generated text features in training process; $\mathcal{L}_{a}$ and $\mathcal{L}_{s}$ refer to two loss functions proposed in section~\ref{Text Generation via Prompt Learning}.
}
\label{table4}
\vspace{4pt}
\begin{tabular}{ccc|ccccc}
\hline
Text & $\mathcal{L}_{a}$ & $\mathcal{L}_{s}$ & A & C & P & R & Avg.\\  
\hline
& & & 86.07 & 74.77 & 90.70 & 90.89 & 85.61\\
$\checkmark$ & $\checkmark$ & &86.81 & 75.78 & 91.33 & 91.35 & 86.32\\
$\checkmark$ & & $\checkmark$ &86.73 & 76.52 & 91.33 & 91.14 & 86.43 \\
$\checkmark$ & $\checkmark$ & $\checkmark$ &86.90 & 77.41 & 91.78 & 91.42 & 86.88\\
\hline
\end{tabular}
\vspace{-20pt}
\end{center}
\end{table}

\setlength{\tabcolsep}{4pt}
\begin{table*}[ht]
\renewcommand\arraystretch{1.2}
\small%
\begin{center}
\caption{Ablation study on single source domain generalization tasks. Similar to section~\ref{Ablation Study of Proposed Modules}, we investigate the effectiveness of proposed modules on PACS, OfficeHome, VLCS and TerraIncognita datasets. For every part of the table, domains in the first row are source domains, and domains in the second row are the remaining target domains.  
}
\label{table5}
\vspace{4pt}
\begin{tabular}{l|ccc|ccc|ccc|ccc|cc}
\hline
\multicolumn{1}{c|}{\multirow{2}{*}{PACS}} & \multicolumn{3}{c|}{A} & \multicolumn{3}{c|}{C} & \multicolumn{3}{c|}{P} & \multicolumn{3}{c|}{S} & \multicolumn{1}{c}{\multirow{2}{*}{Avg.}} & \multicolumn{1}{c}{\multirow{2}{*}{Gain}}\\ 
\cline{2-13}
\multicolumn{1}{c|}{} & C & P & \multicolumn{1}{c|}{S} & A & P & \multicolumn{1}{c|}{S} & A & C & \multicolumn{1}{c|}{S} & A & C & P \\ \hline
ERM  & 96.80 & 99.82 & 93.08 & 95.51 & 99.16 & 79.77 & 91.50 & 97.27 & 86.97 & 91.55 & 96.50  & 85.39 & 92.78 & -\\
\textbf{+Text} & 97.31&  99.70&  90.20 & 94.09&  99.22&  83.58  & 92.48&  97.87& 89.31  & 95.85&  97.31&  97.01 & 94.50 & \textcolor[rgb]{0,0.8,0}{$+1.72$}\\
\textbf{+Norm} & 96.89 & 99.64  & 93.33  & 97.75 &  99.82 &  89.23  & 90.19&  97.70 & 89.39 & 96.97 &  96.20 &  91.08 & 94.85 & \textcolor[rgb]{0,0.8,0}{$+2.07$}\\
\hline
\multicolumn{1}{c|}{\multirow{2}{*}{Office}} & \multicolumn{3}{c|}{A} & \multicolumn{3}{c|}{C} & \multicolumn{3}{c|}{P} & \multicolumn{3}{c|}{R} & \multicolumn{1}{c}{\multirow{2}{*}{Avg.}} & \multicolumn{1}{c}{\multirow{2}{*}{Gain}}\\ 
\cline{2-13}
\multicolumn{1}{c|}{} & C & P & \multicolumn{1}{c|}{R} & A & P & \multicolumn{1}{c|}{R} & A & C & \multicolumn{1}{c|}{R} & A & C & P \\ \hline
ERM & 68.89 & 78.87 & 85.17 & 80.88&  83.58&  83.50 & 74.21&  65.98&  85.38 & 81.58&  6983&  88.42 & 78.86 & -\\
\textbf{+Text} & 71.29&  82.90&  87.26 & 81.38&  86.78&  86.23 & 78.95&  68.18&  88.18 & 84.18&  72.42&  88.02 & 81.31 & \textcolor[rgb]{0,0.8,0}{$+2.45$}\\
\textbf{+Norm} & 70.84&  83.35&  88.02 & 83.52&  86.71&  87.54 & 78.86& 69.44&  88.48 & 84.75&  73.10&  91.01 & 82.14 & \textcolor[rgb]{0,0.8,0}{$+3.28$}\\
\hline
\multicolumn{1}{c|}{\multirow{2}{*}{VLCS}} & \multicolumn{3}{c|}{C} & \multicolumn{3}{c|}{L} & \multicolumn{3}{c|}{S} & \multicolumn{3}{c|}{V} & \multicolumn{1}{c}{\multirow{2}{*}{Avg.}} & \multicolumn{1}{c}{\multirow{2}{*}{Gain}}\\ 
\cline{2-13}
\multicolumn{1}{c|}{} & L & S & \multicolumn{1}{c|}{V} & C & S & \multicolumn{1}{c|}{V} & C & L & \multicolumn{1}{c|}{V} & C & L & S \\ \hline
ERM & 63.67 & 68.19 & 65.23 & 92.01 & 69.74  & 77.46 & 85.23 & 67.13  & 81.75  & 96.75 &  68.64 & 87.54 & 76.95 & -\\
\textbf{+Text} & 65.14&  71.21&  72.99 & 96.25&  72.18&  79.71 & 86.78&  65.40&  81.84  & 96.40&  65.96&  87.48 & 78.45 & \textcolor[rgb]{0,0.8,0}{$+1.50$}\\
\textbf{+Norm} & 66.11&  73.03&  75.36 & 96.82&  75.38&  81.07  & 91.31&  64.98&  85.07 & 97.81&  68.00&  87.96 & 80.24 & \textcolor[rgb]{0,0.8,0}{$+3.29$}\\
\hline
\multicolumn{1}{c|}{\multirow{2}{*}{Terra}} & \multicolumn{3}{c|}{L100} & \multicolumn{3}{c|}{L38} & \multicolumn{3}{c|}{L43} & \multicolumn{3}{c|}{L46} & \multicolumn{1}{c}{\multirow{2}{*}{Avg.}} & \multicolumn{1}{c}{\multirow{2}{*}{Gain}}\\ 
\cline{2-13}
\multicolumn{1}{c|}{} & L38 & L43 & \multicolumn{1}{c|}{L46} & L100 & L43 & \multicolumn{1}{c|}{L46} & L100 & L38 & \multicolumn{1}{c|}{L46} & L100 & L38 & L43 \\ \hline
ERM & 53.68 & 31.69 & 44.37 & 56.95 & 21.06 & 29.19 & 55.07 & 47.52 & 50.33 & 37.21&  29.71&  64.71 & 43.81 & -\\
\textbf{+Text} & 52.36&  35.92&  42.10   & 58.62&  23.22&  26.18 & 48.53&  46.12&  52.12 & 51.57&  39.28& 63.88 & 44.99 & \textcolor[rgb]{0,0.8,0}{$+1.18$}\\
\textbf{+Norm} & 51.82 & 36.73 & 44.89  & 55.81 &  24.03 &  25.17  & 50.79 & 48.18 & 51.22  & 48.01 & 39.70 & 65.06 & 45.26 & \textcolor[rgb]{0,0.8,0}{$+1.45$}\\

\hline
\end{tabular}
\vspace{-16pt}
\end{center}
\end{table*}

\noindent\textbf{Loss Functions in Text Generation Module}
We conduct an ablation study of the text generation module on OfficeHome dataset to verify the necessity of proposed loss functions $\mathcal{L}_s$ and $\mathcal{L}_a$, results are shown in Table~\ref{table4}. We can observe that the performance of the model drops reasonably when any one of the proposed two loss functions is removed.

Without $\mathcal{L}_s$, features extracted from texts containing the same category word but different domain words are aligned to image features in the same category directly, thus diversity between texts containing different domain words is reduced, and text prompt template is trained to be insensitive to the change of domain words. With $\mathcal{L}_a$ decreasing, generated text features provide less valuable domain information, and the generalizability of the model drops accordingly.
Without $\mathcal{L}_a$, features extracted from texts containing different domain words are forced to be as diverse as possible, which could harm the category information in text features. Text features with mistaken category information not only help little to generalization of classifier, but also may weaken its classification ability.

\noindent\textbf{Analysis on Single Source Generalization}
Experiments above are conducted on Leave-one-domain-out generalization tasks, which take multi domains as source domains during training and take a single domain as the target domain during testing. To verify the effectiveness of proposed modules with fewer source domains available during training, we conduct an extra ablation study on single source generalization tasks, in which models are trained on a single source domain and tested on multi target domains. 
Similar to the ablation study above, we compare the performance of baseline ERM, ERM with text modules, and integral TDG containing both the text modules and the normalized classifier. 
Experimental results are shown in Table~\ref{table5}. 

We can see that because the number of source domains reduces, the performance of ERM is remarkably worse compared with its performance in Leave-one-domain-out generalization tasks. Not surprisingly, the generalizability of models is improved by a large margin after adding generated texts into training, and the normalized classifier still performs better than the linear classifier on all of the single source generalization benchmarks. In detail, our method improves the baseline for an average of $2.07\%$ on PACS dataset, $3.28\%$ on OfficeHome dataset,  $3.29\%$ on VLCS dataset, and  $1.45\%$ on TerraIncognita dataset. 
The reason for the experimental results above is obvious: when the domain diversity of source images reduces, texts with abundant domain information are much more helpful to the generalization of models.

\section{Conclusions}
In this paper, we propose a novel method utilizing texts to improve the performance of the language-vision model on domain generalization tasks. 
We first generate diverse domain-relevant words through a lexical substitution model; then we train a text prompt template to create texts with specific information by filling it with different words; finally, we utilize text features extracted from generated texts to improve the generalizability of models.
Experimental results show that the language-vision model with large-scale pre-training has great potential on DG tasks, and it is possible to tap its potential with the help of texts, thus further improving the performance.


{\small
\bibliographystyle{ieee_fullname}
\bibliography{egbib}
}

\end{document}